\documentclass{article} 
\usepackage{iclr2027_conference,times}

\usepackage{amsmath,amsfonts,bm}

\def\eqref#1{equation~\ref{#1}}

\def\1{\bm{1}}

\DeclareMathAlphabet{\mathsfit}{\encodingdefault}{\sfdefault}{m}{sl}
\SetMathAlphabet{\mathsfit}{bold}{\encodingdefault}{\sfdefault}{bx}{n}

\newcommand{\R}{\mathbb{R}}

\definecolor{PastelRedd}{RGB}{216, 27, 96}
\definecolor{cadmiumgreen}{rgb}{0.0, 0.42, 0.24}
\definecolor{customgreen}{rgb}{0.0, 0.4, 0.0}
\definecolor{ours}{rgb}{0.95,0.95,0.95}
\usepackage[
	colorlinks=true,
	linkcolor=PastelRedd,        
	citecolor=brown,  
	filecolor=black,
	urlcolor=blue
]{hyperref}
\usepackage{url}
\usepackage{booktabs}       
\usepackage{graphicx}
\usepackage{subcaption}
\usepackage{arydshln}
\usepackage{amsmath}
\usepackage{amssymb}
\usepackage{amsthm}
\usepackage{mathtools}
\usepackage{amsfonts}       
\usepackage{nicefrac}       
\usepackage{microtype}      
\usepackage{listings}
\usepackage[table]{xcolor}
\usepackage{wrapfig}
\usepackage{cleveref}
\usepackage{multirow}
\usepackage{algorithm}
\usepackage{algorithmic}

\usepackage{tcolorbox}
\usepackage{enumitem}
\newcommand{\method}{Soft-TF}
\newcommand{\gr}[1]{\textcolor{gray}{#1}}

\theoremstyle{plain}

\theoremstyle{remark}

\DeclareMathOperator{\conv}{conv}

\newcommand{\simplex}{\Delta^{|V|-1}}

\theoremstyle{definition}

\title{Soft Teacher Forcing}

\author{Mohammad Fahes\thanks{Work done as student researcher at Google Deepmind.} , Cordelia Schmid, Anurag Arnab \\
Google Deepmind\\
}

\iclrfinalcopy 
\begin{document}

\maketitle  
\lhead{}                            
\renewcommand{\headrulewidth}{0pt} 

\begin{abstract}
Soft tokens have so far been used to replace discrete chain-of-thought in latent reasoning methods. These approaches typically require multi-stage training and, in some cases, auxiliary modules, yet they still underperform supervised fine-tuning (SFT) on $\langle$input, chain-of-thought, output$\rangle$ triples. In this paper, we ask whether soft tokens can be used to improve SFT itself, instead of as an avenue for latent reasoning. We introduce Soft Teacher Forcing (\method{}), a simple training algorithm that combines teacher-forced embeddings with soft tokens through random convex combinations. The soft tokens are computed in a single parallel forward pass over the ground-truth sequence. \method{} adds no parameters, no external model and no curriculum, and leaves inference unchanged. Experiments show that \method{} improves over SFT on GSM8K, MATH, MathQA and AQuA, by +2.1 to +4.1 points on GSM8K across seven Gemma, Llama and Mistral models.
\end{abstract}

\section{Introduction}

Large Language Models (LLMs) have demonstrated remarkable capabilities on complex reasoning tasks through Chain-of-Thought (CoT)~\citep{wei2022chain,wang2023selfconsistency}, in which the model autoregressively generates intermediate tokens that explicitly trace the steps toward a final answer. Building on this, supervised fine-tuning (SFT) on $\langle$\textit{input}, \textit{CoT}, \textit{output}$\rangle$ triples has become a standard approach for training LLMs for reasoning~\citep{mukherjee2023orca,yue2024mammoth,chung2024scaling,shao2024deepseekmath}, before further Reinforcement Learning (RL)~\citep{lightman2024let,liu2024deepseek,team2025kimi}. 
SFT uses teacher forcing (TF)~\citep{jordan1986attractor,williams1989learning}, which conditions each prediction on the preceding ground-truth tokens rather than on the model's own outputs. This allows training to be fully parallelized and aligns naturally with the next-token prediction objective that underlies modern LLMs.

A separate line of work has explored soft tokens, which are continuous inputs such as convex combinations of vocabulary embeddings or last-layer hidden states. The motivation is that a continuous space may be more expressive than a discrete one~\citep{hao2025training,zhang2025soft}, and soft tokens have primarily been used to replace discrete CoT~\citep{deng2024explicit,hao2025training,shen2025codi,xu2025softcot}. However, these latent reasoning methods still underperform the SFT models they use as initialization on real-world mathematical reasoning benchmarks~\citep{cobbe2021training}.
Being recursive by nature, they deviate from TF and are also computationally expensive: generating $n$ latent tokens requires $n+1$ sequential forward passes, and backpropagating through them requires storing every intermediate activation in memory.

While these difficulties are often interpreted as evidence that soft tokens are not yet ready to replace discrete CoT, we view them as an opportunity to ask a different question: rather than attempting to \emph{replace} discrete reasoning, can soft tokens be used to \emph{improve} standard SFT? We answer this positively by introducing \textbf{Soft Teacher Forcing (\method{})}, an algorithm that trains on random convex combinations of teacher-forced (\textit{i.e.}, ground-truth) embeddings and soft tokens. Given
a training sequence, a single forward pass with causal masking over the ground truth yields the model's own next-token distribution at every position. Mapping these distributions back through the embedding matrix produces a soft token per position, in parallel and without any unrolled computation graph. Each input embedding is then replaced by a convex combination of its ground-truth embedding and the corresponding soft token. \method{} introduces no additional parameters, requires no external models or complex training curricula, and leaves both the loss and the inference procedure unchanged.
 
We observe consistent improvements across 4 mathematical reasoning benchmarks and 7 open-source models spanning the Gemma~\citep{team2024gemma,team2024gemma2,team2025gemma3}, Llama~\citep{grattafiori2024llama} and Mistral~\citep{jiang2023mistral7b} families. Moreover, we also outperform latent reasoning methods without the need for external models or multi-stage training. 
Our results establish soft tokens as a highly practical ingredient for supervised post-training, independent of their viability for latent reasoning.

\section{Method}
\label{sec:method}

\subsection{Preliminaries}
\label{sec:prelim}
We first establish the notation used throughout this paper. Let $V$ be a tokenized vocabulary of size $|V|$. Let $S = [q_{1:M},r_{1:N}]$ be an input sequence of length $L=M+N$, comprising $M$ question tokens $q$ and $N$ response tokens $r$. Response tokens include both intermediate CoT steps $c$ and final answer tokens $a$. The subsequence $Y = [r_{1:N}]$ represents the full response. 
$\mathtt{LM}_{\theta}$ is a language model parameterized with $\theta$ acting as a next-token predictor, i.e., $\mathtt{LM}_{\theta}(s_i,s_{<i})$ models the conditional probability for $s_i$ given the preceding prefix $s_{<i}$.
$\mathtt{LM}_\theta$ consists of an input embedding layer $W_{enc}$, a transformer decoder $\Phi$ and a projection layer $W_{dec}$ followed by a softmax operation, meaning $\mathtt{LM}_\theta(\cdot) = \mathtt{Softmax}(W_{dec} \cdot \Phi (W_{enc}(\cdot)))$. 
We write $\delta_v$ for the one-hot vector of token $v$ and $e_v= W_{enc}\delta_v$ for its embedding, such that a sequence $s_{1:L}$ has an embedding matrix $E=[e_{s_1},...,e_{s_L}]$. We index positions in the full sequence by $i \in \{1,\dots,L\}$, meaning response tokens occupy positions $M < i \le L$ and $y_j = s_{M+j}$. We denote by $\simplex := \{q \in \R^{|V|}_{\geq 0} : \mathbf{1}^\top q = 1\}$ the probability simplex over the vocabulary, and by $\mathcal{E} := \{e_v\}_{v \in V} \subset \R^d$ the set of vocabulary embeddings (assumed pairwise distinct), where $d$ is the embedding dimension. We write $\conv(\mathcal{E})$ for the convex hull of $\mathcal{E}$, and note that it is exactly the image of the simplex under the embedding layer: $\conv(\mathcal{E}) = \{W_{enc}q: q \in \simplex \}$.

\textbf{Teacher Forcing (TF).~}
In transformer-based LLMs, TF is the implicit standard. The model receives the entire \textit{ground-truth} sequence at once, and causal masking ensures that the prediction at position $t$ is conditioned solely on the ground-truth tokens from positions $1$ to $t-1$. The next-token prediction objective minimizes the negative log-likelihood of each token in the response sequence $Y$, conditioned on the \textit{ground-truth} prefix rather than the model's own predictions:
\begin{equation}
\label{eq:tf-loss}
\mathcal{L}(\theta) = -\,\mathbb{E}_{\chi} \left[ \sum_{i=1}^{N} \log p_\theta(y_i \mid q_{1:M}, y_{<i}) \right]
\end{equation}
where $\chi$ is the true distribution of sequences $S$. The same objective applies to each token during pretraining. Inference is autoregressive, meaning that generation is token-by-token. Under greedy decoding, this is expressed as:
\begin{equation}
\hat{y}_i = \arg\max_{y \in V} \; p_{\theta}(y \mid q_{1:M}, \hat{y}_{<i})
\end{equation}
Next, we start by explaining the deviation from TF in current methods that use soft tokens.

\subsection{Latent reasoning deviates from teacher forcing}
\label{sec:deviation-tf}
In this section, we present the general framework for training-based continuous latent reasoning methods~\citep{hao2025training,shen2025codi}. In CoCoNut-style latent reasoning~\citep{hao2025training}, the last-layer hidden states are used as continuous tokens. The following formulation is generic and applies to output-based latent reasoning methods. Let $S^{(0)} = [q_{1:M}]$ denote the question prefix, and $E^{(0)} = W_{enc}(S^{(0)})$ the corresponding embeddings. Latent tokens are generated iteratively by extending the embedding sequence and re-running the transformer. For $k \geq 0$, define:
\[
h_M^{(k)} = \left[\Phi(E^{(k)})\right]_{M+k}, \quad
E^{(k+1)} = [E^{(k)}, h_M^{(k)}].
\]
Assuming the total number of latent tokens is $P$, the latent token sequence is:
\[
r_{1:P}^{\text{latent}} = [h_M^{(0)}, h_M^{(1)}, \dots, h_M^{(P-1)}].
\]
Each latent token requires a full forward pass over the extended sequence, leading to a computational cost that grows with both sequence length and $P$. Finally, the full latent sequence is:
\[
E_{\text{latent}} = [E^{(0)}, r_{1:P}^{\text{latent}}, W_{enc}(r_{P+1:N})],
\]
where $r_{P+1:N}$ denotes the remaining discrete response tokens.

The loss function in this case is:
\begin{equation}
\label{eq:coconut_style_loss}
\mathcal{L}(\theta) = -\,\mathbb{E}_{\chi'} \left[ \sum_{i=P+1}^{N} \log p_{\theta}(y_i \mid E^{(0)}, r_{1:P}^{\text{latent}}, E_{r_{P+1:i-1}}) \right]
\end{equation}
where $\chi'$ is the distribution of latent sequences, which has a continuous support. We can see from \Cref{eq:coconut_style_loss}, unlike \Cref{eq:tf-loss}, that only the response part after the latent phase is supervised, and that each token is conditioned on the previous output-based embeddings instead of ground-truth data, which induces a growing deviation from TF. The drift increases with the latent sequence length $P$, which places the LLM in an out-of-distribution (OOD) regime since it was never trained on recursively generated hidden-state trajectories.

\textbf{Compression: a goal or a necessity?} The previous arguments suggest that compression is not merely desirable in latent reasoning models but necessary to mitigate the divergence induced by recursive latent generation. It is also empirically shown that increasing the number of latent tokens significantly harms the performance and leads to training instability~\citep{hao2025training,wei2026simcot}. This is coupled with computational challenges, since gradients propagate through all latent steps, requiring backpropagation through a computation graph unrolled over $P$ iterations.

\begin{figure}[t]
  \vskip 0.2in
  \begin{center}
    \centering
    \includegraphics[width=1.\linewidth]{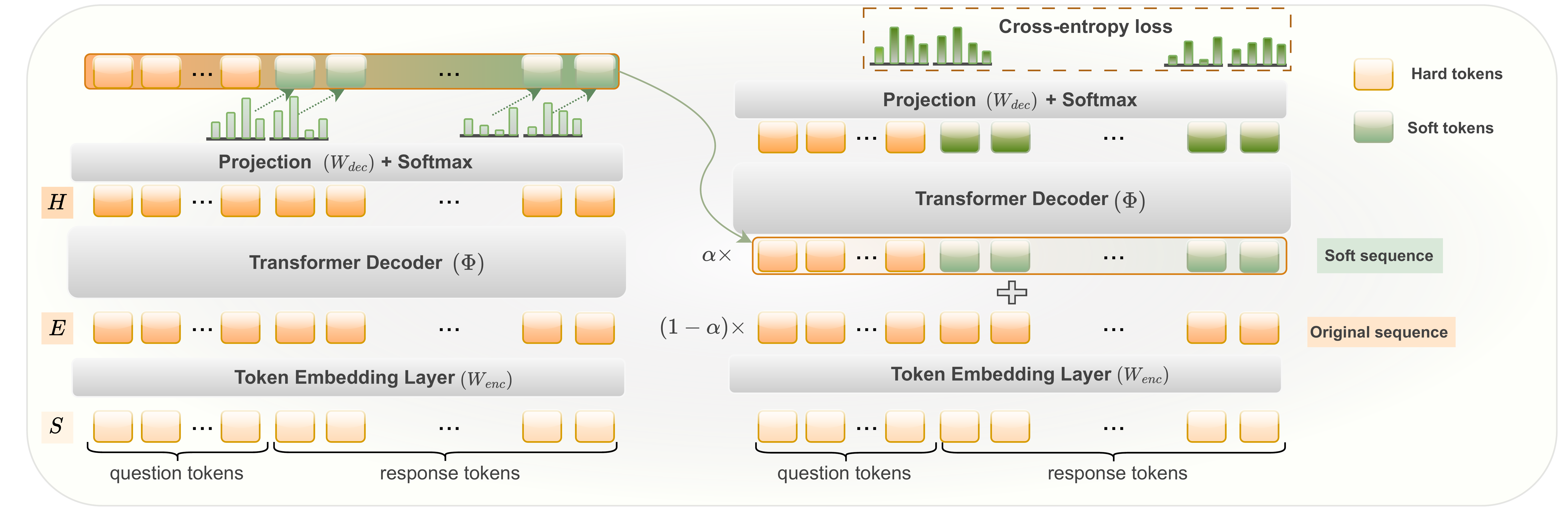}
    \caption{
      (\textit{Left}) \textbf{Teacher-forced soft sequence computation.} Given an initial input sequence, we forward it through the LLM to obtain output probability vectors. These are used as weights for convex combinations of vocabulary tokens, forming soft tokens. (\textit{Right}) \textbf{Training with \method{}.} We train the LLM with the next-token prediction objective on convex combinations of the original sequence embeddings and soft sequence embeddings. We call this framework Soft Teacher Forcing (\method{}).
    }
    \label{fig:main_method}
  \end{center}
\end{figure}

\subsection{Soft Teacher Forcing (\method{})}
\label{sec:soft-tf-sec}
We argue that, unlike latent reasoning methods, the effective use of soft tokens can be unlocked by maintaining the proximity to TF during fine-tuning, otherwise the LLM will be in an out-of-distribution regime. To achieve this, we train LLMs on random convex combinations of teacher-forced embeddings and soft tokens.
\method{} consists of two main steps: \textit{i) Teacher-forced soft sequence computation}, and \textit{ii) training with \method{}}. 

\textbf{Teacher-forced soft sequence computation.~}
Given training sequences of the form $S = [q_{1:M}, r_{1:N}]$, we perform a single parallel forward pass through the LLM to obtain the predicted distribution at every position, $p = [p_1, \dots, p_L] = \mathtt{LM}_\theta(S)$, where $p_i = p_\theta(\cdot \mid s_{\le i})$ is the distribution over the token at position $i+1$. We use these to construct soft tokens at the response positions: for $M < i \le L$,
\begin{equation}
\label{eq:soft-seq-comp}
e^{\texttt{soft}}_i = W_{enc}\,p_{i-1} = W_{enc}\,p_\theta(\cdot \mid s_{<i}) ,
\end{equation}
while $e^{\texttt{soft}}_i = E_i$ at question positions $i \le M$. We also consider a top-$k$ variant, in which $p_{i-1}$ is restricted to its $k$ most probable tokens and renormalized before projection.
Teacher-forced soft sequence computation is illustrated in~\Cref{fig:main_method}~(\textit{Left}). 
Two key properties distinguish this from latent reasoning methods~\citep{hao2025training,shen2025codi}, where soft tokens are generated recursively (each feeding back as input for the next). First, our construction is \textit{teacher-forced}: causal masking ensures that the soft token at position $i$ is conditioned on the ground-truth prefix $s_{<i}$. The recursive construction loses this immediately, as discussed in \Cref{sec:deviation-tf}. 
Second, our construction is \textit{parallel}: all soft tokens are obtained using a single forward pass, requiring only one additional pass regardless of sequence length. Conversely, the recursive alternative demands one pass per soft token and forces all intermediate activations to be retained in memory for backpropagation. We ablate the number of forward passes $K$ in \Cref{sec:ablations} and find that performance degrades as $K$ grows beyond a small number of passes, so the single pass is not only the cheaper option but also more effective. 
We conjecture that this reflects the deviation from the teacher-forcing algorithm used during pretraining, and that remaining close to it matters when fine-tuning.

\begin{wrapfigure}{r}{0.5\linewidth}
\vspace{-10pt}
\centering
\begin{minipage}{\linewidth}
\begin{algorithm}[H]
  \caption{Soft Teacher Forcing (\method{})}
  \label{alg:soft-tf}
  \begin{algorithmic}
    \STATE {\bfseries Input:} original input sequence $S$ $\in \mathbb{N}^{B \times L}$, loss mask $\texttt{Mask}$ (\texttt{True} for response tokens, \texttt{False} else), language model $\mathtt{LM}_\theta$.
    \STATE $E=W_{enc}(S)$
    \STATE $p = \mathtt{stopgrad}\big(\mathtt{Softmax}(W_{dec}\cdot\Phi(E))\big)$
    \STATE ${E^{soft}}= (1-\texttt{Mask}) * E + \texttt{Mask} * W_{enc}p_{\texttt{shift}}$
    \STATE $\alpha \sim \texttt{Beta}(\beta, \beta)$
    \STATE $E^{\texttt{final}} = \alpha E^{\texttt{soft}}+ (1-\alpha) E$
    \STATE $\hat{p} = \mathtt{Softmax}(W_{dec}\cdot\Phi(E^{\texttt{final}}))$
    \STATE $\mathcal{L}= \texttt{Mask} * \texttt{cross-entropy}(\hat{p},p_{\texttt{GT}})$
  \end{algorithmic}
\end{algorithm}
\end{minipage}
\vspace{-10pt}
\end{wrapfigure}
\textbf{Training with \method{}.~} Once the sequence with soft tokens is computed, we train on convex combinations of ground-truth embeddings and soft tokens. This introduces another level of proximity to teacher forcing on top of the teacher-forced soft sequence computation. Specifically, the final sequence for training is computed as:
\begin{equation}
    \label{eq:final-seq}
    E^{\texttt{final}} = \alpha \underbrace{E^{\texttt{soft}}}_{\text{Soft term}} + (1-\alpha) \underbrace{E}_{\text{TF term}},
\end{equation}
where $\alpha~\sim \text{Beta}(\beta,\beta)$ is sampled at each iteration, with hyperparameter $\beta \in (0,\infty)$. This sequence ($E^{\texttt{final}}$) is then forwarded through $\Phi$, and the final probabilities for supervision are computed by projecting the hidden states via $W_{dec}$ followed by \texttt{Softmax} (\textit{cf.}~\Cref{fig:main_method}~(\textit{Right})). \Cref{alg:soft-tf} summarizes the \method{} process.

We highlight that training with $\alpha=0$ reduces \method{} to TF. 
Moreover, we investigate if the second level of proximity to TF is necessary, or if training with $\alpha=1$ suffices.
Experimentally, we show that it is better to train on convex combinations of $E$ and $E^{\texttt{soft}}$, and that randomizing $\alpha$ during training is better than fixing its value.

While \Cref{eq:final-seq} is written as a mixture of two embeddings, $W_{enc}$ is linear, meaning the operation is better understood one layer earlier, i.e., on the simplex. Indeed, for every response position $M < i \le L$ and every $\alpha \in [0,1]$, 
\begin{equation}
\label{final-seq-hull}
    E^{\texttt{final}}_i = W_{enc}\,q_i, 
\end{equation}
where $q_i := \alpha\, p_{i-1} + (1-\alpha)\,\delta_{s_i} \in \simplex$. Consequently $E^{\texttt{final}}_i \in \conv(\mathcal{E})$.

Thus, \method{} does not arbitrarily corrupt an embedding; rather, it feeds the model the embedding of a perturbed distribution. It retains mass $1-\alpha+\alpha\,p_{i-1}(s_i)$ on the ground-truth token and redistributes the remainder according to the model's own beliefs. Training with $\alpha=0$ (\textit{i.e.} $q_i = \delta_{s_i}$) reduces \method{} to TF. Importantly, \method{} ensures that the transformer does not encounter a point outside the convex hull of its vocabulary embeddings. 
This property is broken when last-layer hidden states are used as soft tokens, since they are outputs of $\Phi$ rather than images of $\simplex$ under $W_{enc}$ (i.e., they are not of the form $W_{enc}q$ for any distribution $q$ over the vocabulary).

\subsection{Inference}
\label{sec:inference-sec}

At inference, given an input sequence $Q = [q_1,q_2,...,q_M]$, there are multiple ways to perform generation through autoregression. The first is the (i) \textit{standard} discrete inference, exemplified by greedy decoding where, at each step, the most probable token is used as the next input. The second follows~\citet{zhang2025soft}, where soft tokens are generated until the maximum prediction probability corresponds to an $<$\texttt{end-of-thought}$>$ token. We call this setting (ii) \textit{soft inference}. More generally, we can perform inference using both approaches:
\begin{equation}
    \label{eq:mixed-inference}
    \hat{e}_i = \gamma \hat{e}^{\texttt{soft}}_i + (1-\gamma) \hat{e}^{\texttt{greedy}}_i,
\end{equation}
where $\gamma \in [0, 1]$ is a constant, e.g., $0.5$. We call this setting (iii) \textit{mixed} inference. A fourth type of inference we consider is the (iv) \textit{hybrid} inference, where we perform \textit{mixed} sampling ($\gamma=0.5$) for $K$ steps; $K$ being the number of forward passes used for soft sequence computation---$K=1$ by default---and \textit{standard} greedy decoding after. These strategies are illustrated in Appendix~\ref{sec:additional_illustrations}.

Next, we present experiments with \method{}. Across different inference strategies, we find that standard discrete inference (i.e., $\gamma=0$) works best in most cases (cf.~\Cref{sec:ablations}), so we adopt as a default. This aligns with recent findings by~\citet{butt2026soft}, where soft tokens with added noise are used to train instruction-tuned models with Reinforcement Learning (RL), yet inference is discrete.

\section{Experiments}
\label{sec:experiments}

\subsection{Experimental setup}
\label{sec:experimental-setup}

\textbf{Datasets.~}We evaluate our method mainly on mathematical datasets with reasoning traces, including GSM8K~\citep{cobbe2021training}, MATH~\citep{hendrycks2021measuring}, MathQA~\citep{amini2019mathqa} and AQuA~\citep{ling2017program}. We train on the training set of each dataset and report the accuracy of the test sets. GSM8K, MATH, MathQA, AQuA are composed of $7\,473$, $12\,000$, $29\,837$ and $97\,467$ training samples respectively. The test sets contain $1\,319$, $500$, $2\,985$ and $254$ samples respectively.

\textbf{Implementation details.~} We report main results using the Gemma2-2B~\citep{team2024gemma2} model, but also corroborate our approach with Gemma3-1B, Gemma3-4B, Gemma2-9B, Llama3.2-1B, Llama3.2-3B, and Mistral-7B-v0.3. Since our work is centered on SFT, we use pretrained checkpoints of these models. We run all the experiments with $5$ seeds and report the mean and standard deviation. More implementation details and paired significance test are provided in Appendix~\ref{details-exp-setup}.

\textbf{Prompt formatting.} For all datasets, the question is followed directly by the rationale, and the answer is separated using a $<$\texttt{start-of-answer}$>$ token. For MathQA and AQuA, which are multiple choice questions, the options/choices are inserted between the question and the rationale.

\textbf{Metrics.} At inference, we consider as output all tokens generated after the $<$\texttt{start-of-answer}$>$ token. Exact match accuracy (\%) is used as metric for evaluation. We adopt greedy decoding, i.e., tokens with top-1 probability are considered during autoregressive generation. 

\subsection{Main Results}

\textbf{Baselines.} We compare \method{} to the following baselines:
\begin{itemize}
    \item TF (No-CoT), trained on $<$\textit{question}, \textit{answer}$>$ pairs in a supervised manner using TF.
    \item TF (CoT), trained on $<$\textit{question}, \textit{CoT}, \textit{answer}$>$ triples in a supervised manner using TF.
    \item Label smoothing~\citep{szegedy2016rethinking,pereyra2017regularizing}, a standard output-side regularizer which was first proposed in computer vision then used in neural machine translation~\citep{vaswani2017attention}. Specifically, we train with TF on $<$\textit{question}, \textit{CoT}, \textit{answer}$>$ triples, where the one-hot target $\delta_{y_i}$ is replaced by $(1-\lambda)\delta_{y_i} + \lambda u$ with $u$ uniform over $V$. This is the output-side counterpart of \method{}, which softens the input instead.
    \item Scheduled sampling~\citep{mihaylova2019scheduled} is a method proposed to mitigate exposure bias, i.e., the mismatch between conditioning on ground-truth previous tokens during training and on model-generated tokens during inference. During training, scheduled sampling randomly replaces ground-truth previous tokens with model-generated tokens according to a schedule that controls the probability of using generated tokens.
    \item Original CoCoNut~\citep{hao2025training}, a latent reasoning approach that progressively removes the explicit CoT, and trains the model to perform the corresponding reasoning in latent space.
    We divide the tokenized CoT into $3$ parts and remove one part per stage of training. 
    \item 
    A modified variant CoCoNut$^\dagger$ without multi-stage training nor compression, where reasoning is trained using hidden states but supervised explicitly with CoT. This baseline isolates the effect of the compression procedure in latent reasoning. More details are in Appendix~\ref{details-baselines}.
    \item SoftThinking which uses soft tokens only during inference, similar to \citep{zhang2025soft}. Here we use the TF (CoT) model.
    \item SoftThinking$^\dagger$, where we extend SoftThinking by applying it during training.
\end{itemize}

\begin{table}[t]
  \caption{Performance (accuracy \%) of \method{} compared against baselines on GSM8K, MATH, MathQA and AQuA datasets. All methods fine-tune Gemma2-2B pretrained checkpoint, except SoftThinking, which is an inference-only method using the TF (CoT) checkpoint. 
  For CoCoNut$^\dagger$, we report the best decoding strategy per dataset (see Appendix~\ref{sec:coconut-dagger}).}
  \label{main-results}
  \begin{center}
    \begin{small}
      \begin{sc}
       \resizebox{0.99\textwidth}{!}{
        \begin{tabular}{lcccc}
          \toprule
          Method  & GSM8K  & MATH & MathQA & AQuA \\
          \midrule
          TF (no CoT)   & 14.00$\pm$ 0.28 & 12.40$\pm$ 0.32 & 49.92$\pm$ 0.49 & 40.71$\pm$ 1.94  \\
          TF (CoT) & 44.70$\pm$ 0.89 & 14.44$\pm$ 0.91 & 59.86$\pm$ 0.80 & 38.19$\pm$ 4.30   \\
          \midrule
          \gr{CoCoNut (Stage 1)} & \gr{26.03$\pm$ 0.69} & \gr{9.32$\pm$ 1.07} & \gr{46.55$\pm$ 1.01} & \gr{32.13$\pm$ 2.06} \\
          \gr{CoCoNut (Stage 2)} & \gr{19.36$\pm$ 0.78} & \gr{6.80$\pm$ 0.47} & \gr{48.17$\pm$ 0.88} & \gr{31.18$\pm$ 0.85} \\
          \gr{CoCoNut}~\gr{\citep{hao2025training}} & \gr{13.59$\pm$ 0.56} & \gr{10.08$\pm$ 1.01} & \gr{45.16$\pm$ 1.12} & \gr{34.96$\pm$ 1.98} \\
          \gr{CoCoNut$^\dagger$}~\gr{\citep{hao2025training}} & \gr{21.00$\pm$ 1.25} & \gr{0.92$\pm$ 0.77} & \gr{12.14$\pm$ 2.28} & \gr{12.05$\pm$ 2.58} \\
          \midrule
          SoftThinking~\citep{zhang2025soft} & 41.62$\pm$ 1.21 & 11.44$\pm$ 0.77 & 57.71$\pm$ 0.43 & 36.69$\pm$ 4.00 \\
          SoftThinking$^\dagger$~\citep{zhang2025soft} & 44.70$\pm$ 1.50 & 13.12$\pm$ 0.98 & 54.69$\pm$ 0.79 & 37.64$\pm$ 3.27 \\
          Label smoothing & 41.53$\pm$0.37 & 15.16$\pm$1.05 & 59.73$\pm$0.87 & 39.05$\pm$2.20 \\
          Scheduled sampling~\citep{mihaylova2019scheduled} & 46.38$\pm$1.02 & 15.04$\pm$1.76 & 59.78$\pm$1.01 & 39.45$\pm$2.32\\
          \method{} (Ours) & \textbf{49.28}$\pm$ 0.59 &  \textbf{16.20}$\pm$ 1.33 & \textbf{61.49}$\pm$ 0.96 &  \textbf{42.06}$\pm$ 1.95\\
          \bottomrule
        \end{tabular}
        }
      \end{sc}
    \end{small}
  \end{center}
  \vskip -0.1in
\end{table}

Illustrations of CoCoNut$^\dagger$ and SoftThinking$^\dagger$ are provided in Appendix~\ref{sec:additional_illustrations}. 
Results in~\Cref{main-results} show that \method{} is the only method that consistently outperforms TF (CoT) across all datasets. 

We observe that CoCoNut not only underperforms TF (CoT) but also TF (No-CoT), despite using TF (CoT) weights as initialization. These results are in line with~\citet{hao2025training} showing limitations of CoCoNut for model sizes beyond GPT-2~\citep{radford2019language} ($\sim$ 117M parameters). Additionally, by projecting the hidden states and decoding them, we found that the corresponding top-1 predictions are not interpretable and exhibit repetitive patterns. More details are provided in Appendix~\ref{sec:decoding-latent}.

CoCoNut$^\dagger$ removes the compression and multi-stage training, obtaining soft tokens from a single forward pass but using hidden states instead of convex combinations of vocabulary embeddings (with $\alpha=1$ in~\Cref{eq:final-seq}). It improves over CoCoNut on GSM8K but is much worse elsewhere, which supports prior findings that latent reasoning requires compression~\citep{hao2025training,wei2026simcot}. Results with different sampling strategies are in Appendix~\ref{sec:coconut-dagger}.

SoftThinking~\citep{zhang2025soft} is applied during inference using the TF (CoT) model, providing additional evidence to \citet{wu2026llms} that vanilla convex reasoning does not lead to improvements. We also evaluate SoftThinking$^\dagger$, a variant that applies SoftThinking during training: \Cref{eq:soft-seq-comp} remains unchanged, while $\alpha=1$ in~\Cref{eq:final-seq}. This variant shows improvement over SoftThinking, hinting at the potential of using soft tokens for training. However, it underperforms \method{}, highlighting the importance of training with a convex combination of soft tokens and teacher-forced embeddings (i.e., random $\alpha$ in~\Cref{eq:final-seq}). This variant therefore serves as a direct ablation of the mixing step: it matches the TF (CoT) baseline on GSM8K ($44.70$), and the $+4.58$ gain appears only once the soft sequence is
mixed with the teacher-forced one.

\textbf{Softening the target instead of the input.} As discussed in \Cref{sec:soft-tf-sec}, \method{} replaces the one-hot input $\delta_{s_i}$ by a mixture of the ground-truth token and the model's own prediction, which is the input-side analogue of label smoothing on the output. Label smoothing improves over TF (CoT) on MATH ($15.16$ against $14.44$) and AQuA ($39.05$ against $38.19$) but not on GSM8K ($41.53$ against $44.70$) or MathQA ($59.73$ against $59.86$), whereas \method{} improves on all four datasets. Softening the target is therefore not sufficient, and the gain does not come from smoothing per se.

\textbf{Comparison to scheduled sampling.} Empirically, scheduled sampling~\citep{bengio2015scheduled,mihaylova2019scheduled} has been shown to improve only marginally and occasionally upon teacher forcing. In our experiments, scheduled sampling performs generally better than label smoothing, but still lags behind \method{} on all four datasets. Since both methods use the same top-$5$ soft tokens (hyperparameters in Appendix~\ref{details-baselines}), this comparison isolates the effect of interpolating rather than replacing ground-truth embeddings, and whether a remplacement schedule is required. The Bernoulli ablation in \Cref{tab:alpha-distribution-ablation}, which replaces each embedding by its soft token with a fixed probability of $0.5$, points the same way: both replacement schemes fall below \method{}.

\begin{table}[!t]
\centering
\caption{Accuracy (\%) on the GSM8K dataset using different models.}
\label{other-versions}
\begin{small}
\begin{sc}
\resizebox{0.7\linewidth}{!}{%
\begin{tabular}{lccc}
  \toprule
  Method & TF (no CoT) & TF (CoT) & \method{} (Ours) \\
  \midrule
  Gemma3-1B & 6.82$\pm$0.42 & 16.77$\pm$0.50 & \textbf{19.38}$\pm$0.60 {\tiny\textbf{\textcolor{customgreen}{$\uparrow +2.61$}}} \\
  Gemma2-2B & 14.00$\pm$0.28 & 44.70$\pm$0.89 & \textbf{48.52}$\pm$1.24 {\tiny\textbf{\textcolor{customgreen}{$\uparrow +3.82$}}} \\
  Gemma3-4B & 18.54$\pm$0.48 & 54.86$\pm$0.50 & \textbf{57.95}$\pm$1.23 {\tiny\textbf{\textcolor{customgreen}{$\uparrow +3.09$}}} \\
  Gemma2-9B & 30.37$\pm$0.44 & 73.89$\pm$0.77 & \textbf{77.07}$\pm$1.19 {\tiny\textbf{\textcolor{customgreen}{$\uparrow +3.18$}}} \\
  Llama3.2-1B & 5.90$\pm$0.26 & 28.23$\pm$0.88 & \textbf{30.33}$\pm$0.59 {\tiny\textbf{\textcolor{customgreen}{$\uparrow +2.10$}}} \\
  Llama3.2-3B & 13.18$\pm$0.26 & 48.58$\pm$1.20 & \textbf{52.66}$\pm$1.12 {\tiny\textbf{\textcolor{customgreen}{$\uparrow +4.08$}}} \\
  Mistral-7B-v0.3 & 13.91$\pm$0.95 & 57.57$\pm$0.80 & \textbf{60.82}$\pm$0.66 {\tiny\textbf{\textcolor{customgreen}{$\uparrow +3.25$}}} \\
  \bottomrule
\end{tabular}}
\end{sc}
\end{small}
\end{table}

\textbf{\method{} with other model sizes.} We train \method{} using Gemma3-1B, Gemma3-4B and Gemma2-9B, Llama3.2-1B, Llama3.2-3B, and Mistral-7B-v0.3 models.~\Cref{other-versions} shows that \method{} consistently leads to improvements across different LLMs.

\subsection{Design Choices \& Ablations}
\label{sec:ablations}
We ablate \method{} design choices ($\alpha$, soft token type, and computation) using Gemma2-2B on GSM8K unless noted otherwise.

\begin{wraptable}{r}{0.5\linewidth}
\vspace{-10pt}
\centering
\caption{Ablation study on soft tokens in \method{}. 
}
\begin{small}
\begin{tabular}{lcc}
  \toprule
  Method  & GSM8K & MATH \\
  \midrule    
  \textcolor{gray}{TF (CoT)} & \textcolor{gray}{44.70$\pm$ 0.89} & \textcolor{gray}{14.44$\pm$ 0.91}  \\
  \midrule
  hidden & 46.81$\pm$ 1.11 & 14.40$\pm$ 0.94 \\
  argmax & 48.35$\pm$ 1.06 & 13.76$\pm$ 1.11 \\
  convex-full & 48.52$\pm$ 1.24 & 15.48$\pm$ 1.50  \\
  convex-top-5 & \textbf{49.28}$\pm$ 0.59 & \textbf{16.20}$\pm$ 1.33\\
  \bottomrule
\end{tabular}
\end{small}

\label{cnvx-vs-hidden}
\vspace{-10pt}
\end{wraptable}
\textbf{Which soft tokens are better?~} Prior work builds soft tokens either as convex combinations of vocabulary embeddings~\citep{zhang2025soft,wu2026llms,butt2026soft} or from last-layer hidden states~\citep{hao2025training,shen2025codi}. We test the latter within our framework, which means using the last-layer hidden states at position $i-1$ in place of $W_{enc}p_{i-1}$.
\Cref{cnvx-vs-hidden} shows that hidden states underperform convex soft tokens and do not improve over TF (CoT) on MATH. 
We also compare against argmax (top-1) tokens, which replace $p_{i-1}$ with $\delta_{\hat y}$ in~\Cref{eq:soft-seq-comp}. This variant remains a convex combination of two vocabulary embeddings; only the direction of the perturbation differs, pointing towards the single most likely token rather than towards the full predictive distribution. On GSM8K the two are close ($48.35$ against $49.28$), while on MATH the argmax variant falls below the TF (CoT) baseline ($13.76$ against $14.44$) and soft tokens reach $16.20$.

\begin{wrapfigure}{r}{0.31\linewidth}
    \centering
    \includegraphics[width=\linewidth]{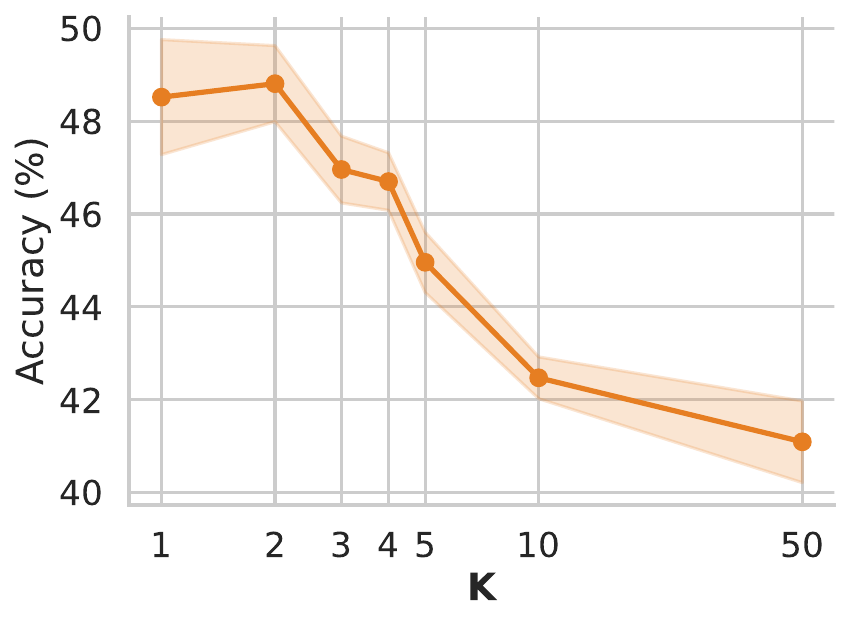}
    \caption{Ablation on the number of forward passes $K$ for soft sequence computation. The x-axis is on a log scale.}
    \label{fig:sequential-vs-parallel}
\end{wrapfigure}

\textbf{Teacher-forced \textit{vs} sequential soft tokens.~} Soft sequence computation is teacher-forced, i.e., performed in a single forward pass through the model. As discussed in \Cref{sec:soft-tf-sec}, this keeps every soft token conditioned on the ground-truth prefix and avoids multiple forward passes. To test this choice, we instead compute soft tokens sequentially, feeding each one back as input to produce the next:
\begin{equation}
\label{seq-ssc}
e_{r_k} = W_{enc}\,\mathtt{Softmax}\big(W_{dec}\,\Phi(E^{(0)}, e_{r_1}, \dots, e_{r_{k-1}})\big),
\end{equation}
where $e_{r_1} = W_{enc}\,p_\theta(\cdot \mid q_{1:M})$. $E^{(0)} = W_{enc}(q_{1:M})$ and $\Phi$ attends over the question and all previous soft tokens, its output being taken at the last position. 
This resembles CoCoNut~\citep{hao2025training} except that soft tokens are convex combinations of
vocabulary embeddings rather than last-layer hidden states. The number of forward passes $K$ sets the length of the sequential dependency, and $K = 1$ recovers teacher-forced soft sequence computation.
Increasing $K$ removes the teacher-forcing signal from a growing part of the input.
\Cref{fig:sequential-vs-parallel} shows that performance degrades as $K$ grows beyond a small number of passes. We illustrate sequential computation in Appendix~\ref{sec:additional_illustrations}.

\textbf{Random vs. constant $\alpha$.} In this experiment, we investigate whether it is necessary to use a randomly sampled $\alpha$ during training. Thus, we apply \method{} training using constant values of $\alpha$ ranging from $0$ to $1$ in increments of $0.1$. \Cref{tab:constant-alpha} shows that performance significantly lags behind that obtained using continuously sampled random $\alpha$ values during training.

\textbf{$\alpha$ sampling distribution.~}
By default, $\alpha$ (\Cref{eq:final-seq}) is sampled from $\mathrm{Beta}(0.1,0.1)$, a U-shaped distribution concentrated near 0 and 1. We compare it to $\mathcal{U}[0,1]$ and to $\mathrm{Bernoulli}(0.5)$, which $\mathrm{Beta}(0.1,0.1)$ approximates continuously, where each embedding is either ground-truth or soft token. 
As shown in \Cref{tab:alpha-distribution-ablation}, Beta outperforms the uniform law with both full and top-5 soft tokens, and also outperforms Bernoulli, more significantly on MATH than GSM8K.
Performance is thus non-monotone in the interior mass on $[0,1]$, which we report as an empirical finding.

\begin{table}[t]
\newcommand{\varh}[1]{} 
\caption{\textbf{Choice of $\alpha$.} (\subref{tab:constant-alpha}) Constant \textit{vs.} random $\alpha$: \method{} with constant $\alpha \in ]0,1]$. The dashed horizontal line shows accuracy (\%) with $\alpha\sim\text{Beta}(0.1,0.1)$ as a reference. Shaded regions show the standard deviation over 5 runs. (\subref{tab:alpha-distribution-ablation}) Sampling distribution of $\alpha$: Beta$(0.1, 0.1)$ vs. Uniform$[0, 1]$ vs. Bernoulli$(0.5)$ with $\alpha \in \{0, 1\}$, with the TF (CoT) baseline as reference.}

\setlength{\fboxsep}{1pt}
\centering

\begin{subtable}[t]{0.45\linewidth}
    \vspace{0pt} 
    \centering
    \includegraphics[width=0.8\linewidth]{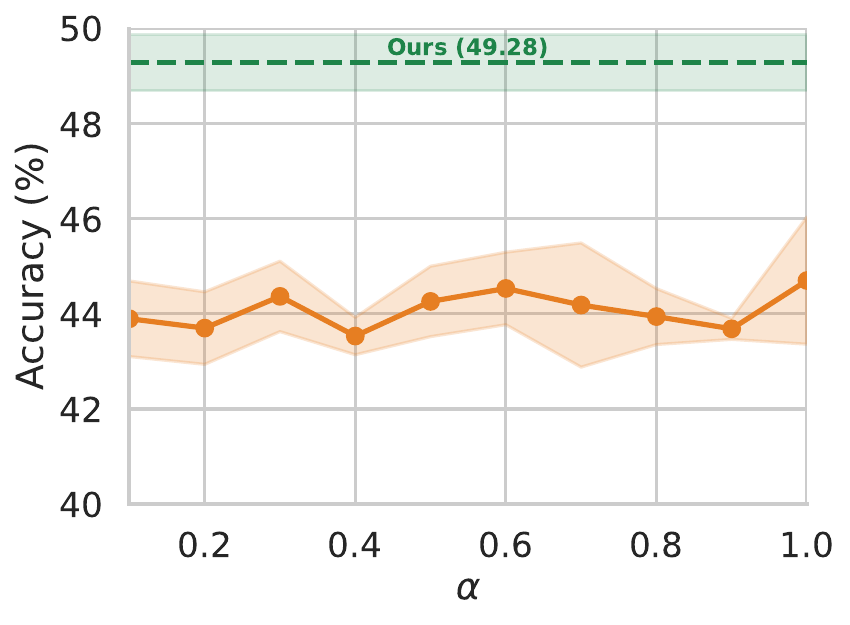}
    \vspace{-10pt}
    \caption{Constant \textit{vs.} random $\alpha$}
    \label{tab:constant-alpha}
\end{subtable}%
\hfill%
\begin{subtable}[t]{0.55\linewidth}
    \vspace{8pt}
    \centering
    \resizebox{0.7\linewidth}{!}{%
        \setlength{\tabcolsep}{0.3em}
        \begin{tabular}{lccc}
            \toprule
              & Method  & GSM8K & MATH \\
            \midrule
              & \textcolor{gray}{TF (CoT)} & \textcolor{gray}{44.70$\pm$ 0.89} & \textcolor{gray}{14.44$\pm$ 0.91}  \\
            \midrule
            \multirow{3}{*}{Full} & Beta & \textbf{48.52}$\pm$ 1.24  & \textbf{15.48}$\pm$ 1.50 \\
              & Uniform & 47.61$\pm$ 1.04 & 15.12$\pm$ 0.94 \\
              & Bernoulli & 47.92$\pm$ 1.01 & 13.52$\pm$ 0.97 \\
            \midrule
            \multirow{3}{*}{top-5} & Beta & \textbf{49.28}$\pm$ 0.59 & \textbf{16.20}$\pm$ 1.33 \\
              & Uniform & 47.52$\pm$0.66 & 13.60$\pm$ 2.02 \\
              & Bernoulli & 48.63$\pm$ 0.60 & 13.44$\pm$ 1.40 \\
            \bottomrule
        \end{tabular}
    }%
    \caption{Sampling distribution of $\alpha$}
    \label{tab:alpha-distribution-ablation}
\end{subtable}%
\label{tab:xxxx}
\end{table}

\begin{wraptable}{r}{0.55\linewidth}
  \vspace{-1em}
  \caption{Accuracy (\%) on the GSM8K dataset for different sampling strategies.}
  \label{main-results-gsm8k-all-sampling-strategies}
  \vspace{0.5em}
  \centering
  \small
  \resizebox{\linewidth}{!}{%
    \begin{sc}
      \begin{tabular}{lccc}
        \toprule
        Method & TF (no CoT) & TF (CoT) & \method{} (Ours) \\
        \midrule
        Standard ($\gamma=0$)  & 14.00$\pm$ 0.28 & 44.70$\pm$ 0.89 & \textbf{48.52}$\pm$ 1.24 \\
        Soft ($\gamma=1$)      & 13.62$\pm$ 0.41 & 41.62$\pm$ 1.21 & \textbf{43.31}$\pm$ 0.84 \\
        Mixed ($\gamma=0.5$)   & 14.10$\pm$ 0.26 & 43.99$\pm$ 1.05 & \textbf{46.50}$\pm$ 1.00 \\
        Hybrid                 & 14.06$\pm$ 0.27 & 44.49$\pm$ 1.29 & \textbf{48.72}$\pm$ 1.46 \\
        \bottomrule
      \end{tabular}
    \end{sc}
  }
  \vspace{-1em}
\end{wraptable}
\textbf{Sampling strategies.} As previously mentioned in~\Cref{eq:mixed-inference}, inference can be performed using discrete tokens, soft tokens, or a combination of the two. Here, we present four sampling strategies and show that there is no clear advantage to using soft tokens during inference. The inference strategies are as follows: (\textit{i}) Discrete greedy decoding, which we call \textit{Standard}. This corresponds to $\gamma=0$ in~\Cref{eq:mixed-inference} and is the default strategy used throughout the paper. (\textit{ii}) \textit{Soft} sampling, which corresponds to $\gamma=1$. (\textit{iii}) \textit{Mixed} sampling, which corresponds to $\gamma=0.5$ and consists of mixing greedy decoding embeddings with soft embeddings. (\textit{iv}) \textit{Hybrid} sampling, which uses \textit{Mixed} sampling ($\gamma=0.5$) for $K$ steps and \textit{Standard} greedy decoding thereafter. This strategy is motivated by the fact that, to mimic training, mixed sampling should be performed for the same number of forward steps used for soft sequence computation during training, and greedy decoding for the rest. 
Results in~\Cref{main-results-gsm8k-all-sampling-strategies} demonstrate that \textit{standard} sampling works well within our framework, with hybrid sampling yielding a slight improvement at the cost of higher standard deviation.

\begin{wraptable}{r}{0.55\linewidth}
\vspace{-10pt}
  \caption{Comparison to Noisy TF. Accuracy (\%) is reported across different noise levels $\sigma$.}
  \label{tab:noisy-tf}
  \begin{center}
    \begin{small}
      \begin{sc}
      \resizebox{0.9\linewidth}{!}{%
        \begin{tabular}{lccc}
          \toprule
          \multicolumn{2}{c}{Method} & GSM8K & MATH \\
          \midrule    
          \multicolumn{2}{c}{TF (CoT)} & 44.70$\pm$ 0.89 & 14.44$\pm$ 0.91  \\
          \midrule
          \multirow{10}{*}{Noisy TF (CoT) 
          } & Noise level ($\sigma$) \\
          \cdashline{2-2} \\
          & 0.1 & 44.50$\pm$ 1.13 & 14.04$\pm$ 1.02 \\
          & 0.3 & 45.23$\pm$ 0.30 & 14.44$\pm$ 0.41 \\
          & 0.5 & 45.44$\pm$ 0.65 & 14.12$\pm$ 0.84 \\
          & 0.7 & 45.55$\pm$ 1.03 & 13.72$\pm$ 1.79\\
          & 0.9 & 45.00$\pm$ 1.73 & 14.16$\pm$ 1.05 \\
          & 2.0 & 43.00$\pm$ 0.69 & 12.96$\pm$ 1.28 \\
          & 5.0 & 19.97$\pm$ 0.16 & 3.56$\pm$ 0.61 \\ 
          \midrule
          \multicolumn{2}{c}{\method{} (Ours)} & 48.52$\pm$ 1.24 & 15.48$\pm$ 1.50 \\
          \multicolumn{2}{c}{\method{} (Ours) (top-5)} & \textbf{49.28}$\pm$ 0.59 & \textbf{16.20}$\pm$ 1.33 \\
          \bottomrule
        \end{tabular}
    }
      \end{sc}
    \end{small}
  \end{center}
  \vskip -0.1in
\end{wraptable}
\textbf{Comparison to Noisy TF.} Does \method{} only have a regularization effect from adding noise to TF embeddings? To examine this, we compare \method{} to Noisy TF (CoT), a baseline inspired by \citep{jainneftune}, where we add noise to token embeddings during training. Specifically, given an input embedding $e_i$, we augment it with Gaussian noise $e_i^{\text{noisy}} := e_i + \sigma \boldsymbol{\epsilon}$, where $\boldsymbol{\epsilon} \sim \mathcal{N}(\mathbf{0}, \mathbf{I})$ and $\sigma$ is a scalar controlling the noise level.
Results in \Cref{tab:noisy-tf} show that, despite slightly improving accuracy for some values of $\sigma$ on GSM8K, Noisy TF (CoT) lags well behind \method{} at every noise level, indicating that the effect of using soft tokens cannot be replaced by noise.

\section{Related Works}
\label{sec:related}

\textbf{Soft tokens for latent reasoning.~} Soft tokens have been used to replace discrete chain-of-thought with reasoning in a continuous space. iCoT~\citep{deng2024explicit} starts from a model fine-tuned on explicit CoT and gradually removes CoT tokens through a curriculum, internalizing the reasoning into the weights. CoCoNut~\citep{hao2025training} follows the same recipe but replaces
each removed step with last-layer hidden states fed back as input, and CODI~\citep{shen2025codi} self-distills an explicit-CoT teacher into such a latent student by aligning hidden activations. SoftCoT~\citep{xu2025softcot} generates instance-conditioned soft tokens with an auxiliary language model and a trainable projection, while \citet{su2025token} map CoT tokens to discrete latent codes with a VQ-VAE. SoftThinking~\citep{zhang2025soft} instead applies convex combinations of vocabulary embeddings at inference only; \citet{wu2026llms} show that this largely collapses to greedy decoding and that stochasticity is needed to recover a benefit. Two limitations recur across this line of work. First, the methods rely on multi-stage curricula, distillation or auxiliary modules, and still underperform the SFT baseline they build on for real-world mathematical reasoning~\citep{cobbe2021training}. Second, latent reasoning requires compression: adding latent tokens degrades accuracy and destabilizes training, with latents collapsing to a narrow set of decoded tokens~\citep{hao2025training,wei2026simcot} (which we also observe, see Appendix~\ref{sec:decoding-latent}). 
Closest to our setting, \citet{butt2026soft} train instruction-tuned models with RL on noisy soft tokens that replace the discrete CoT in model-generated rollouts, which requires sequential generation, and decode discretely at inference. In contrast, \method{} operates in the supervised setting: soft tokens are computed in a single parallel pass over the ground-truth CoT, are mixed with the ground-truth embeddings rather than replacing them, and are conditioned on the ground-truth prefix. The two approaches are complementary: \method{} could serve as the SFT stage before such RL training.

\textbf{Soft tokens to improve SFT.~} We use soft tokens not to replace the CoT but to improve supervised fine-tuning on it. As in
SoftThinking~\citep{zhang2025soft}, our soft tokens are convex combinations of vocabulary embeddings, but they are used only during training, computed in a single teacher-forced forward pass, and mixed with rather than replacing ground-truth embeddings. \method{} thus requires no recursive generation, compression, auxiliary module, distillation or curriculum, and inference remains standard discrete decoding.
It also differs from mixup~\citep{zhang2018mixup,sun2020mixup}, which interpolates two examples and their labels for data augmentation: \method{} mixes each token's ground-truth embedding with the model's own prediction and leaves the target unchanged.

\textbf{Scheduled sampling.~}Our work is reminiscent of scheduled sampling~\citep{bengio2015scheduled}, which reduces the mismatch between teacher-forced training and autoregressive inference by randomly feeding the model either ground-truth tokens or its own predictions. Closest to ours is \citet{mihaylova2019scheduled}, who adapt it to transformers using two decoder passes and feed the second pass either argmax predictions or mixtures of embeddings weighted by the predicted distribution, following a schedule that decreases the probability of choosing the ground-truth tokens across training. \method{} uses the same soft tokens but, rather than replacing ground-truth embeddings according to a schedule, interpolates between the two with a coefficient drawn from a fixed distribution, so the ground-truth token (i.e., the teacher forcing part) remains partly present at nearly every position. 

\section{Conclusion}
We present Soft Teacher Forcing (\method{}), a training algorithm that leverages soft tokens to improve SFT of LLMs. 
\method{} consists of two components. The first is teacher-forced soft sequence computation, which performs a single forward pass through the model to obtain output probabilities used to compute soft tokens. The second is training on convex combinations of teacher-forced and soft tokens. Extensive experiments demonstrate the effectiveness of our framework. We hope \method{} will help rethink training LLMs with soft tokens. Future work may explore applying \method{} to  
other stages of language model training (e.g. pretraining).

\textbf{Limitations.} \method{} requires one additional forward pass per training step, costing $+42\%$ wall-clock time and $+14\%$ peak memory (\Cref{tab:compute}), a single extra pass regardless of sequence length, so the relative overhead does not grow with the number of soft tokens, while inference cost is unchanged. Our evaluation is confined to mathematical reasoning datasets, and whether \method{} improves SFT in other domains, such as instruction following or code generation, remains untested. All models are fine-tuned with LoRA and have at most $9$B parameters.

\bibliography{iclr2027_conference}
\bibliographystyle{iclr2027_conference}

\appendix
\newpage

\section{Additional experimental details}
\label{details-exp-setup}

\textbf{Implementation details.} 
%
All models are fine-tuned with low-rank adaptation (LoRA)~\citep{hu2022lora} of rank $64$, a batch size of $8$, and a cosine learning-rate schedule with linear warmup over the first $10\%$ of training iterations. We use Adafactor~\citep{shazeer2018adafactor} for Gemma models and AdamW~\citep{loshchilov2018decoupled} for Llama and Mistral models. The peak learning rate is $10^{-2}$ for all Gemma models, $2\times10^{-4}$ for Llama3.2-1B, $10^{-4}$ for Llama3.2-3B, and $5\times10^{-5}$ for Mistral-7B-v0.3. These values are tuned for the TF (CoT) baseline and reused unchanged for \method{}, so any benefit from learning-rate tuning goes to the baseline rather than to our method.

\textbf{Training budget and model selection.}
The purpose of our comparison is to determine whether \method{} yields a real improvement over standard SFT, rather than an improvement that results from a particular training configuration. To this end, we aim to get the most out of every method, including the baselines, and compare each method at its best.
Indeed, methods with different optimization dynamics reach their best performance after different numbers of iterations, and we want to avoid an under-tuned baseline. We thus treat the training budget as a hyperparameter to be tuned for each method, and tune it in exactly the same way for all methods.

Concretely, we sweep the total number of iterations for each method, using the same candidate sets for all methods. Each candidate is a separate run with its own cosine learning-rate schedule, whose warmup and decay are scaled to the total number of iterations. Each run is evaluated at its final iteration; we use no early stopping and no intermediate checkpoint selection. For each method, we report the candidate budget with the highest mean accuracy over $5$ seeds. The results in the main paper are therefore best-versus-best comparisons: each method is evaluated at its own optimal budget, within the same search space and under the same selection rule.

Interestingly, we found that all baselines achieve their best results for the same total number of iterations in most cases. We use $5{,}000$ iterations for GSM8K and MATH, $30{,}000$ for MathQA, $50{,}000$ for AQuA, and $10{,}000$ iterations on Llama3.2-1B/GSM8K. They are identical for TF (CoT) and \method{} in the main experiments except in two cases: (i) on Gemma3-1B/GSM8K, TF (CoT) selects $5{,}000$ iterations whereas \method{} selects $10{,}000$; and (iii) on Gemma2-2B/MATH with top-$5$ soft tokens, TF (CoT) selects $5{,}000$ iterations whereas \method{} selects $10{,}000$.

These cases illustrate why the best-versus-best comparison is the appropriate one: fixing a common budget would favor whichever method prefers that budget. On Gemma3-1B/GSM8K, fixing the budget at the value selected by \method{} ($10{,}000$ iterations) would give a gain of $+3.75$ points, fixing it at $15{,}000$ would give $+0.89$, and fixing it at the value selected by TF (CoT) ($5{,}000$) would give $-0.53$. The best-versus-best gain we report, $+2.61$, compares each method at its own optimum. \Cref{tab:budget-sweep} shows that TF (CoT) cannot achieve \method{} performance (19.38\%) at any training budget.

\begin{table}[h]
\centering
\caption{Gemma3-1B on GSM8K: accuracy (\%) at different total iteration budgets ($5$ seeds). Each method is trained separately at each budget using its corresponding learning-rate schedule and evaluated at the final iteration. Bold marks the budget selected for each method.}
\label{tab:budget-sweep}
\begin{small}
\begin{sc}
\begin{tabular}{lccc}
\toprule
Iterations & TF (no CoT) & TF (CoT) & \method{} \\
\midrule
5k & \textbf{6.82}$\pm$0.42 & \textbf{16.77}$\pm$0.50 & 16.24$\pm$0.70 \\
10k & 6.07$\pm$0.46 & 15.63$\pm$0.74 & \textbf{19.38}$\pm$0.60 \\
15k & 5.70$\pm$0.66 & 16.41$\pm$0.83 & 17.30$\pm$1.00 \\
\bottomrule
\end{tabular}
\end{sc}
\end{small}
\end{table}

\textbf{\method{} hyperparameters.}
The mixing coefficient $\alpha$ (\Cref{eq:final-seq}) is sampled from $\mathrm{Beta}(\beta,\beta)$ with $\beta=0.1$~\citep{zhou2021domain}. It is sampled independently per token on GSM8K and AQuA, and once per sequence on MATH and MathQA. All reported results are means and standard deviations over $5$ seeds, measured at the final training iteration.

\textbf{CoCoNut.} CoCoNut is trained for $5$ epochs per stage, and is initialized from TF (CoT) weights. The CoT is tokenized and divided into three equal chunks, each representing a CoT ``step''. At each stage, one step is removed following \citet{hao2025training}, and replaced with $2$ latent tokens. Each stage is trained with $5$ random seeds, and the best-performing checkpoint is used to initialize the next stage.

\textbf{Prompt formatting.} We provide details on prompt formatting for Gemma by showing examples from each training dataset:

\begin{tcolorbox}[colback=green!1, colframe=gray]
\underline{GSM8K}:
\small
\begin{verbatim}
<start_of_turn>user
Question:
Natalia sold clips to 48 of her friends in April, and then she sold
half as many clips in May. How many clips did Natalia sell altogether 
in April and May?
Answer:
<end_of_turn>
<start_of_turn>model
Natalia sold 48/2 = <<48/2=24>>24 clips in May.
Natalia sold 48+24 = <<48+24=72>>72 clips altogether in April and May.
#### 72<end_of_turn>
\end{verbatim}
\end{tcolorbox}

\begin{tcolorbox}[colback=green!1, colframe=gray]
\underline{MATH}:
\small
\begin{verbatim}
<start_of_turn>user
How many vertical asymptotes does the graph of $y=\frac{2}{x^2+x-6}$ 
have? 
<end_of_turn>
<start_of_turn>model
The denominator of the rational function factors into $x^2+x-6=(x-2)
(x+3)$. Since the numerator is always nonzero, there is a vertical
asymptote whenever the denominator is $0$, which occurs for $x = 2$ 
and $x=-3$. Therefore, the graph has $\boxed{2}$ vertical asymptotes.
#### 2<end_of_turn>
\end{verbatim}
\end{tcolorbox}

\begin{tcolorbox}[colback=green!1, colframe=gray]
\underline{MATHQA}:
\small
\begin{verbatim}
<start_of_turn>user
there are 10 girls and 20 boys in a classroom . what is the ratio of
girls to boys ?
a ) 1 / 2 , b ) 1 / 3 , c ) 1 / 5 , d ) 10 / 30 , e ) 2 / 5 
<end_of_turn>
<start_of_turn>model
if girls is 10 and boys is 20 , then 10 / 20 . so ratio of girls
to boys is = 10 / 20 = 1 / 2 answer : a
#### a<end_of_turn>
\end{verbatim}
\end{tcolorbox}

\begin{tcolorbox}[colback=green!1, colframe=gray]
\underline{AQuA}:
\small
\begin{verbatim}
<start_of_turn>user
Question: Two friends plan to walk along a 43-km trail, starting at
opposite ends of the trail at the same time. If Friend P's rate is 15%
faster than Friend Q's, how many kilometers will Friend P have walked 
when they pass each other?
Choices:
A)21
B)21.5
C)22
D)22.5
E)23
Answer:<end_of_turn>
<start_of_turn>model
If Q complete x kilometers, then P completes 1.15x kilometers.
x + 1.15x = 43
2.15x=43
x = 43/2.15 = 20
Then P will have have walked 1.15*20=23 km.
The answer is E.
#### E<end_of_turn>
\end{verbatim}
\end{tcolorbox}

In all datasets, \#\#\#\# is used as the $<$start\_of\_answer$>$ token.

\begin{wraptable}{r}{0.48\textwidth}
\vspace{-12pt}
\centering
\caption{Measured training cost of \method{} against TF (CoT) on Gemma2-2B, GSM8K.}
\label{tab:compute}
\begin{small}
\begin{sc}
\setlength{\tabcolsep}{3pt}
\resizebox{\linewidth}{!}{%
\begin{tabular}{lccc}
\toprule
Method & Sec/step & Peak mem/device & Throughput \\
\midrule
TF (CoT) & 0.882 & 21.42 GiB & 9.07 samp/s \\
\method{} & 1.255 & 24.47 GiB & 6.38 samp/s \\
\midrule
Overhead & $+42\%$ & $+14\%$ & $0.70\times$ \\
\bottomrule
\end{tabular}}
\end{sc}
\end{small}
\vspace{-10pt}
\end{wraptable}

\textbf{Training cost.}  
\method{} adds one forward pass per training step. 
Since gradients do not backpropagate through the first forward pass, its activations need not be stored, incurring no additional backward cost. \Cref{tab:compute} reports measured wall-clock and peak memory under identical conditions (Gemma2-2B, GSM8K, $4\times$V100, $300$ steps, batch size $8$). The overhead is $+42\%$ in time and $+14\%$ in memory, for an accuracy gain up to $4.58$ percentage points on GSM8K. This contrasts with continuous-CoT methods~\citep{hao2025training}, whose cost grows with the number of latent tokens $P$ ($P{+}1$ forward passes with backpropagation through an unrolled graph). Inference cost is identical to SFT, since \method{} decodes discretely.

\textbf{Paired significance tests.} All comparisons in \Cref{main-results} use $5$ seeds. \Cref{tab:significance} reports two-sided paired $t$-tests, pairing by seed, on the per-seed difference between \method{} and TF (CoT). The improvement is significant on GSM8K, MATH and AQuA. On MathQA, \method{} is higher on all five seeds (differences $+3.35$, $+1.47$, $+0.57$, $+2.68$, $+0.07$) but the test is borderline.

\begin{table}[h]
\centering
\caption{Two-sided paired $t$-tests over $5$ seeds for \method{} against TF (CoT) on Gemma2-2B.}
\label{tab:significance}
\begin{small}
\begin{sc}
\begin{tabular}{lccccc}
\toprule
Dataset & TF (CoT) & \method{} & $\Delta$ & $p$ & 95\% CI \\
\midrule
GSM8K  & 44.70$\pm$0.89 & 49.28$\pm$0.59 & $+4.58$ & 0.0002 & [$+3.56$, $+5.60$] \\
MATH   & 14.44$\pm$0.91 & 16.20$\pm$1.33 & $+1.76$ & 0.0053 & [$+0.88$, $+2.64$] \\
MathQA & 59.86$\pm$0.80 & 61.49$\pm$0.96 & $+1.63$ & 0.058  & [$-0.09$, $+3.34$] \\
AQuA   & 38.19$\pm$4.30 & 42.06$\pm$1.95 & $+3.87$ & 0.0405 & [$+0.27$, $+7.47$] \\
\bottomrule
\end{tabular}
\end{sc}
\end{small}
\end{table}

These tests should be read with caution. With $n=5$ the $t$-test has four degrees of freedom, so the estimates of the standard error, and hence the $p$-values and confidence intervals, are themselves noisy, and the normality assumption cannot be checked at this sample size. This problem applies to all four rows, including those with small $p$-values, and not only to the borderline MathQA comparison. We report the tests because the per-seed differences are consistently positive across datasets, not as a precise quantification of the effect size.

\section{Details about baselines}
\label{details-baselines}

\textbf{Baseline hyperparameters.} For label smoothing, we use $\lambda = 0.1$. For scheduled sampling, the second decoder pass receives mixtures of the top-$5$ predicted token embeddings weighted by their renormalized probabilities, and the probability of feeding the ground-truth token decays linearly from $1$ to $0.5$ over training. These settings performed best in \citet{mihaylova2019scheduled}.

\subsection{Decoding the latent CoT tokens}
\label{sec:decoding-latent}

As demonstrated in \Cref{main-results}, CoCoNut collapses and underperforms the TF (No CoT) baseline. We present some examples of CoCoNut inference here. By projecting the hidden states used for latent reasoning at inference time and decoding them, we find that the corresponding top-1 predictions are not interpretable and exhibit repetitive patterns. This aligns with recent findings in~\citet{wei2026simcot}, which show that failed latent reasoning models generate only a narrow range of decoded tokens, a phenomenon they call ``semantic homogenization''. We observe the same pattern. Below are examples from the GSM8K test set:

\small
\begin{lstlisting}
Prompt: <start_of_turn>user
Sally received the following scores on her math quizzes: 50, 80, 80, 
60, 40, 90, 100, 70, and 60. Find her mean score.<end_of_turn>
<start_of_turn>model
(*@\colorbox{yellow}{Text from latent: The \#\#\#\# \#\#\#\# \#\#\#\# \#\#\#\# \#\#\#\#}@*) 
Prediction: #### 72<end_of_turn>
Ground truth: #### 70<end_of_turn>
----------------------------------------------------------------------
Prompt: <start_of_turn>user
Tim enters a competition and has to try and guess the number of red 
jelly beans in a jar. The jar has a square base and is 6 inches by 6 
inches and 15 inches tall. He knows that a jelly bean is .15 cubic 
inches. He also knows that they get about 80% efficiency in packing.  
He estimates  30% of the jelly beans are red. The true number of red 
jelly beans in the container is 900. How far off was his guess?
<end_of_turn>
<start_of_turn>model
(*@\colorbox{yellow}{Text from latent: \#\#\#\# \#\#\#\# \#\#\#\# \#\#\#\# \#\#\#\# \#\#\#\#}@*)
Prediction: #### 30<end_of_turn>
Ground truth: #### 36<end_of_turn>
----------------------------------------------------------------------
Prompt: <start_of_turn>user
To heat during the winter, Ali ordered 850 kilos of coal. The coal is
delivered in 50 kg bags, costing $18 per unit. How much does his order
cost? <end_of_turn>
<start_of_turn>model
(*@\colorbox{yellow}{Text from latent: \#\#\#\# \#\#\#\# \#\#\#\# \#\#\#\# \#\#\#\# \#\#\#\#}@*)
Prediction: #### 252<end_of_turn>
Ground truth: #### 306<end_of_turn>
\end{lstlisting} 

Examples from the MATH test set:
\small{
\begin{lstlisting}
Prompt: <start_of_turn>user
Let $z = 1+i$ and $w = \dfrac{3z+1}{5z+7}$. Find $|w|$.
<end_of_turn>
<start_of_turn>model
(*@\colorbox{yellow}{Text from latent: We \$ \#\#\#\# \#\#\#\# \#\#\#\# \#\#\#\#}@*)
Prediction: #### \dfrac{\sqrt{10}}{5}<end_of_turn>
\textbf{Ground truth}: #### \frac{5}{13}<end_of_turn>
----------------------------------------------------------------------
Prompt: <start_of_turn>user
Let $n$ be a positive integer.  Simplify the expression
\[\frac{(2^4 + \frac{1}{4})(4^4 + \frac{1}{4}) \dotsm [(2n)^4 + 
\frac{1}{4}]}{(1^4 + \frac{1}{4})(3^4 + \frac{1}{4}) \dotsm [(2n - 1)
^4 + \frac{1}{4}]}.\]
<end_of_turn>
<start_of_turn>model
(*@\colorbox{yellow}{Text from latent: The all \#\#\#\# \#\#\#\# \#\#\#\# \#\#\#\#}@*)
Prediction: #### 2n + 1<end_of_turn>
Ground truth: #### 8n^2 + 4n + 1<end_of_turn>
\end{lstlisting}
}

\subsection{CoCoNut$^\dagger$ results with different sampling strategies.}
\label{sec:coconut-dagger}
\normalsize
In \Cref{main-results}, we include the results of CoCoNut$^\dagger$, a variant of CoCoNut without multi-stage training or compression. Here, we report in \Cref{CoCoNut-variant-different-sampling} the results of this variant for each sampling strategy described in~\Cref{sec:inference-sec}. This experiment supports the idea that compression is necessary for latent reasoning methods and highlights a fundamental problem in using hidden states \textit{alone} for reasoning (\textit{i.e.}, $\alpha=1$ in \Cref{eq:final-seq}).

\begin{table}
  \caption{Performance (accuracy \%) of CoCoNut$^\dagger$ with different sampling strategies on GSM8K, MATH, MathQA, and AQuA datasets. Mixed and soft decoding yield identical results on all four datasets.} 
  \label{CoCoNut-variant-different-sampling}
  \begin{center}
    \begin{small}
      \begin{sc}
      \resizebox{0.9\textwidth}{!}{
        \begin{tabular}{lcccc}
          \toprule
          Method  & GSM8K  & MATH & MathQA & AQuA \\
          \midrule
          TF (no CoT)   & 14.00$\pm$ 0.28 & 12.40$\pm$ 0.32 & 49.92$\pm$ 0.49 & 40.71$\pm$ 1.94  \\
          TF (CoT) & 44.70$\pm$ 0.89 & 14.44$\pm$ 0.91 & 59.86$\pm$ 0.80 & 38.19$\pm$ 4.30   \\
          \midrule
          CoCoNut$^\dagger$ (hybrid) & 20.68$\pm$ 1.71 & 0.52$\pm$ 0.44 & 0.96$\pm$ 0.51 & 11.89$\pm$ 4.48 \\
          CoCoNut$^\dagger$ (standard ($\gamma=0$)) & 21.00$\pm$1.25 & 0.92$\pm$ 0.77 & 0.94$\pm$ 0.69 & 11.18$\pm$ 3.81  \\
          CoCoNut$^\dagger$ (Mixed ($\gamma=0.5$)) & 1.00$\pm$0.38 & 0.04$\pm$ 0.09 & 12.14$\pm$ 2.28 & 12.05$\pm$ 2.58  \\
          CoCoNut$^\dagger$ (soft ($\gamma=1$)) & 1.00$\pm$0.38 & 0.04$\pm$ 0.09 & 12.14$\pm$ 2.28 & 12.05$\pm$ 2.58 \\
          \bottomrule
        \end{tabular}
        }
      \end{sc}
    \end{small}
  \end{center}
  \vskip -0.1in
\end{table}

\begin{figure}
  \vskip 0.2in
  \begin{center}
    \centering
    \includegraphics[width=1.\linewidth]{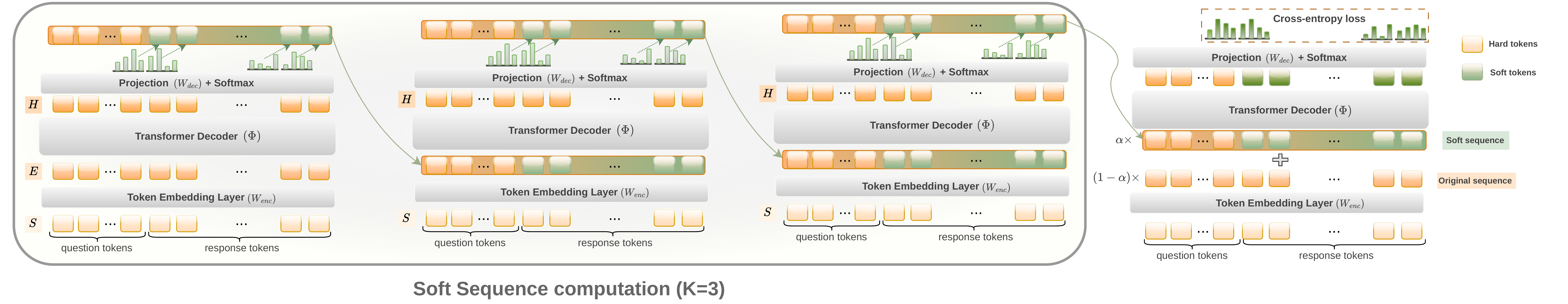}
    \caption{
      Sequential soft sequence computation with $K=3$. In contrast to the teacher-forced soft sequence computation ($K=1$, \textit{cf.} \Cref{fig:main_method}) adopted in this paper, sequential soft sequence computation requires multiple forward passes through the transformer. $K$ reflects the number of soft tokens exhibiting sequential dependency.
    }
    \label{fig:sequential_k_3}
  \end{center}
\end{figure}

\begin{figure}
    \centering

    \begin{subfigure}{0.49\textwidth}
        \centering
        \includegraphics[width=\linewidth]{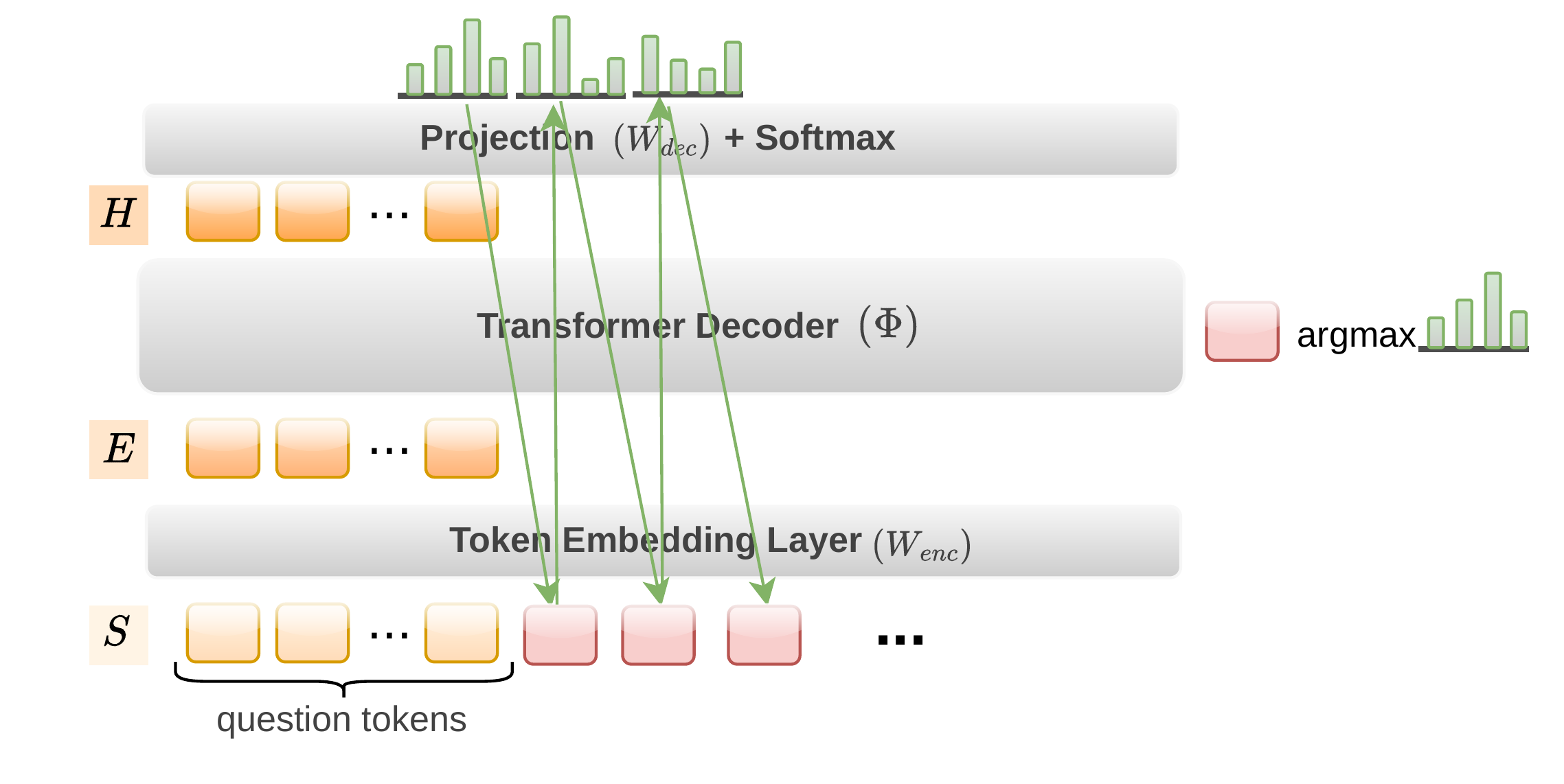}
        \caption{Standard}
    \end{subfigure}
    \hfill
    \begin{subfigure}{0.49\textwidth}
        \centering
        \includegraphics[width=\linewidth]{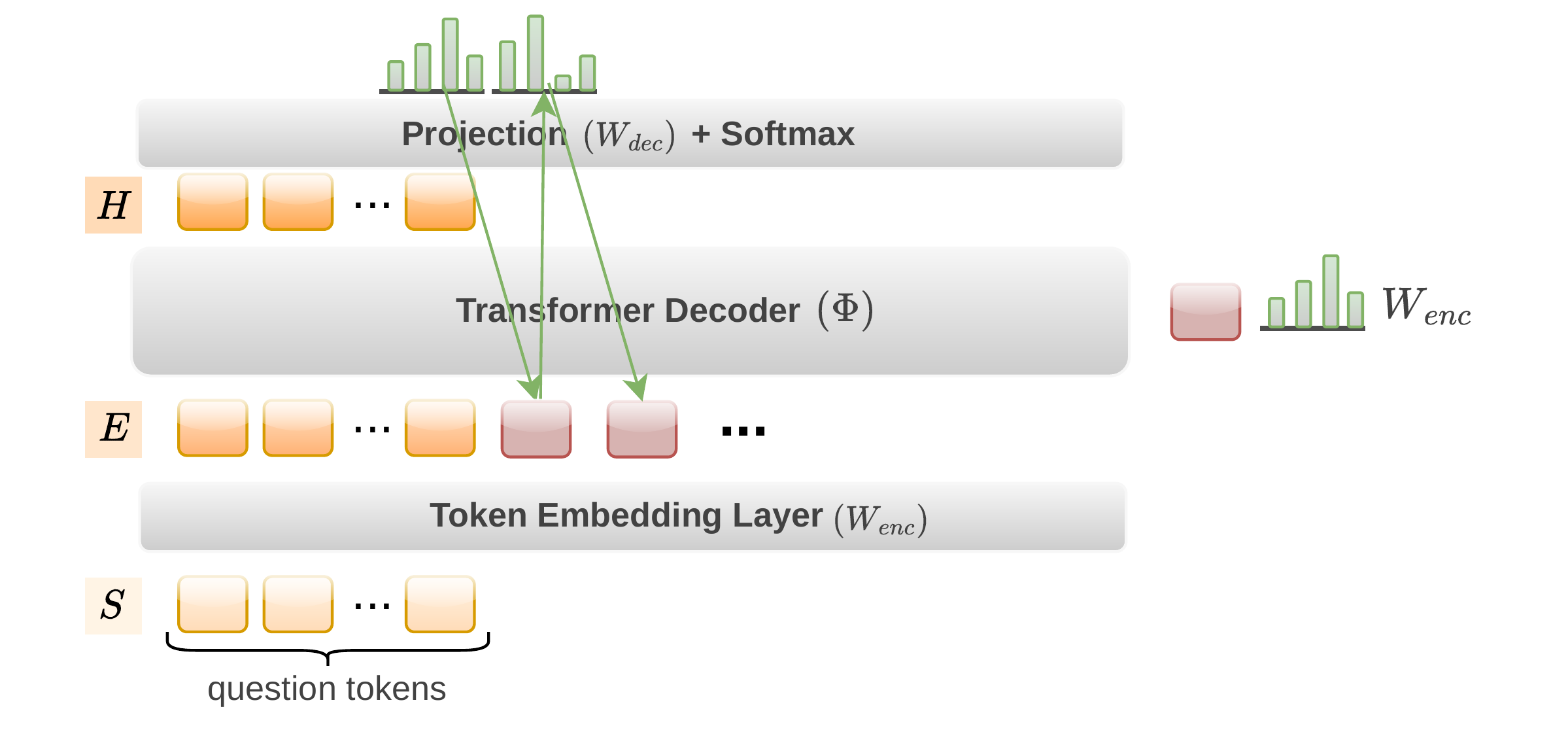}
        \caption{Soft}
    \end{subfigure}

    \vspace{0.5cm}

    \begin{subfigure}{0.49\textwidth}
        \centering
        \includegraphics[width=\linewidth]{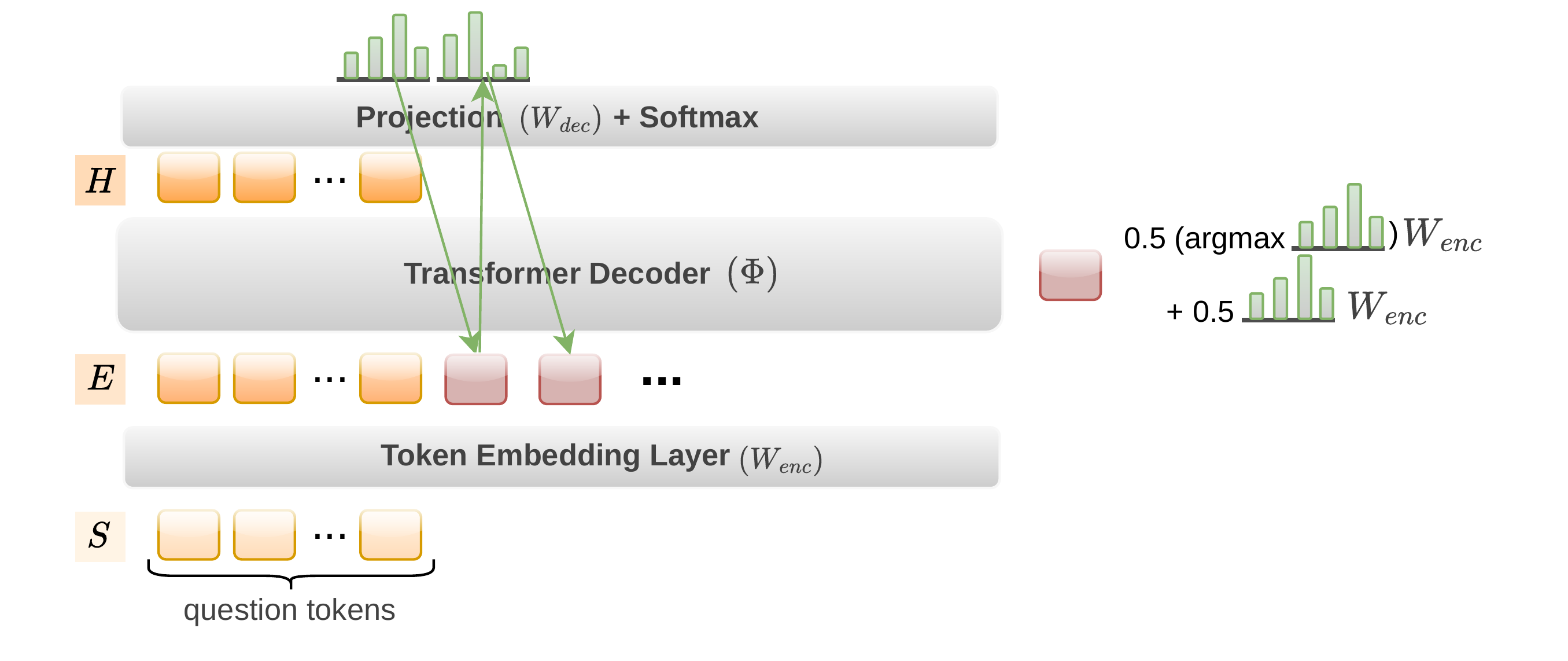}
        \caption{Mixed}
    \end{subfigure}
    \hfill
    \begin{subfigure}{0.49\textwidth}
        \centering
        \includegraphics[width=\linewidth]{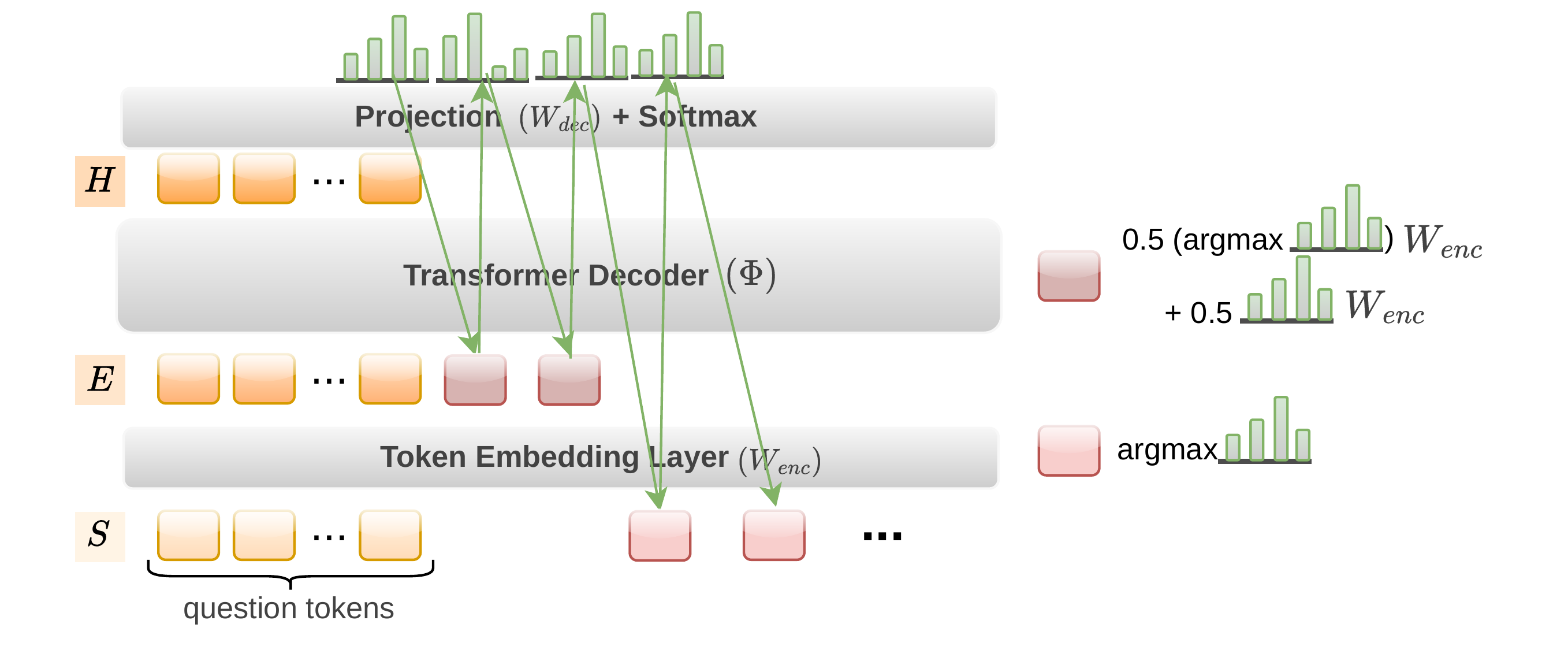}
        \caption{Hybrid ($K=2$)}
    \end{subfigure}

    \caption{Inference strategies described in \Cref{sec:inference-sec}. (a) Standard discrete inference, (b) Soft inference, (c) Mixed inference, (d) Hybrid inference. Note that, for illustration only, we use $K=2$ for hybrid inference, while the ablation in \Cref{main-results-gsm8k-all-sampling-strategies} uses $K=1$.}
    \label{fig:inference-strategies}
\end{figure}

\begin{figure}
  \vskip 0.2in
  \begin{center}
    \centering
    \includegraphics[width=0.9\linewidth]{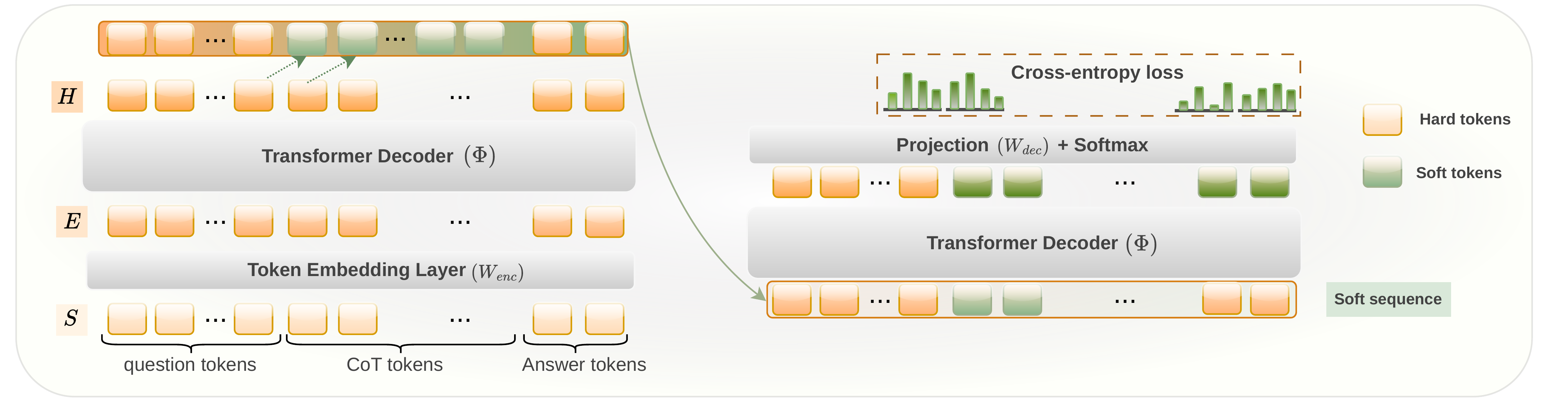}
    \caption{The CoCoNut$^\dagger$ variant, which does not perform compression. Hidden states are used to construct the \textit{teacher-forced} soft sequence, which is then used for training without resorting to the original sequence.
    }
    \label{fig:coconut_dagger}
  \end{center}
\end{figure}

\begin{figure}
  \vskip 0.2in
  \begin{center}
    \centering
    \includegraphics[width=0.9\linewidth]{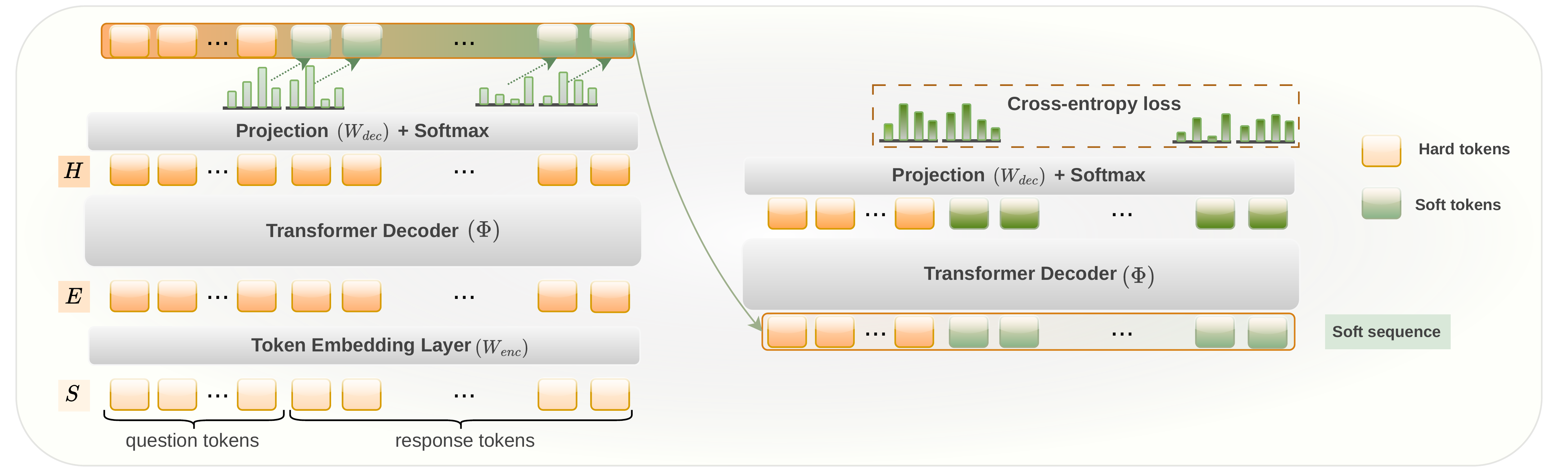}
    \caption{The SoftThinking$^\dagger$ variant extends the original SoftThinking~\citep{zhang2025soft} to training. The \textit{teacher-forced} soft sequence is used for training without resorting to the original sequence.
    }
    \label{fig:softthinking_dagger}
  \end{center}
\end{figure}

\section{Additional illustrations}
\label{sec:additional_illustrations}

To clarify the variants we test, we provide additional illustrations. 

\textbf{Sequential soft sequence computation.} In \Cref{sec:ablations}, we present an ablation study contrasting teacher-forced soft-sequence computation ($K=1$) with sequential computation. In the latter case, $K$ determines the number of latent tokens that exhibit sequential dependency. \Cref{fig:sequential_k_3} illustrates the case where $K=3$.

\textbf{Inference strategies.} In \Cref{sec:inference-sec}, we present four inference strategies and ablate them in \Cref{main-results-gsm8k-all-sampling-strategies}. We illustrate these strategies in \Cref{fig:inference-strategies}.

\textbf{CoCoNut$^\dagger$ and SoftThinking$^\dagger$.} CoCoNut$^\dagger$ and SoftThinking$^\dagger$ are variants of CoCoNut~\citep{hao2025training} and SoftThinking~\citep{zhang2025soft}, respectively. The goal is to adapt these baselines to our setting, in which no compression or multi-stage training is used for CoCoNut, and SoftThinking is applied for both training and inference. CoCoNut$^\dagger$ and SoftThinking$^\dagger$ are illustrated in \Cref{fig:coconut_dagger,fig:softthinking_dagger}, respectively. We note that for CoCoNut$^\dagger$, we use soft tokens only for the CoT part, as in CoCoNut. We also empirically find that CoCoNut$^\dagger$ fails to generate an answer when soft sequence computation involves the answer part.

\end{document}